# Efficient classification using parallel and scalable compressed model and Its application on intrusion detection

Tieming Chen[1*], Xu Zhang[1], Shichao Jin[2], Okhee Kim[2]

[1]College of Computer Science and Technology, Zhejiang University of Technology, Hangzhou, 310023, China

[2]School of Software Microelectronics, Peking University, Beijing, 100190, China

*tmchen@zjut.edu.cn

**Abstract.** In order to achieve high efficiency of classification in intrusion detection, a compressed model is proposed in this paper which combines horizontal compression with vertical compression. OneR is utilized as horizontal compression for attribute reduction, and affinity propagation is employed as vertical compression to select small representative exemplars from large training data. As to be able to computationally compress the larger volume of training data with scalability, MapReduce based parallelization approach is then implemented and evaluated for each step of the model compression process abovementioned, on which common but efficient classification methods can be directly used. Experimental application study on two publicly available datasets of intrusion detection, KDD99 and CMDC2012, demonstrates that the classification using the compressed model proposed can effectively speed up the detection procedure at up to 184 times, most importantly at the cost of a minimal accuracy difference with less than 1% on average.

**Keywords:** Compressed model, MapReduce, Parallelization, Classification, Intrusion Detection.

## 1. Introduction

With the larger and larger amount of network communication data generated, the design of Intrusion Detection System (IDS) with high efficiency has become much more challenging. It is very important to discover abnormal behaviors at early stage, therefore, compared to the traditional signature-based detection, research on anomaly detection has been more popular in academia, as it has the potential power to detect unknown attacks by kinds of heuristic learning on the historical training data.

Anomaly detection generally includes two steps, building a model on training data and using the model for detection. However, training data are usually in a large scale, which can severely impede the detection since many detection models may need to scan all of them in certain cases. Intuitively, an effective and direct way to reduce time cost for detection is to minimize the volume of a model that is used in the detection process, but building a systematic and scalable solution on generating such minimizing training data model for efficient intrusion detection is still in challenge.

To address this problem, our work will pay attention largely to the building of data compression instead of the detection phase, striving to boost the detection efficiency based on a proposed compressed model of training data. Therefore, the solution presented in this paper is applicable for those model-based anomaly detection approaches, especially for the classification based system[1] because extracting the classification model directly from the huge volume of training data instances inevitably needs intense computation.

As for how to build compressed model, our proposal is made through inspecting into the following common natures of the training data, that is to say the motivation and inspiration of our works are generated from the following observations:

a. By analyzing the attributes of the training data, we can easily find that the values of some attributes (features) in the whole training data only range in a small scale, which may have less impact on the detection accuracy.
b. Some training instances are similar, because they are only different from each other on several attributes and the values of these attributes are slightly different, which may have redundancy for detection model building.
c. For high dimensional training data, computing the similarity or some akin metric between each pair of instance is time-consuming. That means, for some novel but promising data processing algorithms such like Affinity Propagation[2], the general computing memory would likely explode when the dimensionality of training data matrix increases to some large extent.

For the purpose of effectively and efficiently handling these problems, we will propose a new framework of compressed model on training data. The model compression procedure mainly includes horizontal compression and vertical compression. The overall idea is presented in Fig. 1, where the first step is to normalize the original data followed by the horizontal compression and the vertical compression sequentially. Based on the compressed model, efficient classification can be directly built to detect new data without losing accuracy.

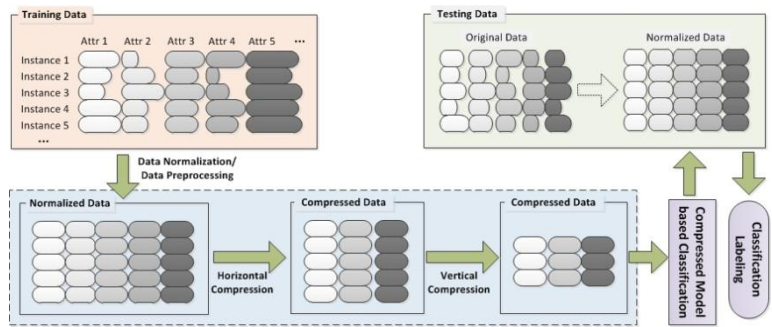

Fig.1 The main idea of compressed model for data classification

Furthermore, to computationally compress as large-scale as the training dataset could be, a cloud-based computing framework will be employed to parallelize the compression procedure, which realizes the scalability on training data compressing.

To summarize, we mainly make the following contributions in this paper:

a. We propose a compressed detection model, which is a compact version from the original training dataset with regard to reducing both feature dimension and instance volume.
b. We implement a MapReduce based parallel computing solution for the abovementioned model compression, which can compress larger scale of training data if only scaling up the involving distributed nodes.
c. We finalize our compressed model-based classification approaches, and demonstrate the performance of our method on detection efficiency and accuracy on two publicly available intrusion detection datasets, KDD99[3] and CDMC2012[4].

The remainder of this paper is organized as follows. Literature works are firstly studied in section 2. Section 3 introduces the methodology of our parallel and scalable compressed model, and explains its implementation procedure in detail. Section 4 describes the efficient classification deployments using the compressed model. Application experiments and its performance analysis on intrusion detection are presented in Section 5, while concluding remark and future work are discussed in last section.

## 2. Related Work

During the past two decades, researchers in related fields have paid much attention to intrusion detection. Signature based detection (e.g. Snort [5]) relies on the knowledge of system vulnerabilities and known attack patterns, which hence is unable to detect unknown attacks. Correspondingly, anomaly detection is more dynamic and be able to detect novel attacks, which has attracted a lot of works worldwide. Generally, an anomaly-based intrusion detection system includes following steps, data gathering, data preprocessing[6], model building[7], and model-based detection[8]. Although some common classification methods can be well used for the model-based detection, the way of model building may affect the detection results directly and heavily. Therefore, academic research on the model-based classification approach for intrusion detection is one unceasing focused topic. At beginning, detection accuracy, usually known as detection rate(recall) and false alarm rate(false positive), is widely concerned for the real-world application purpose. Recently, detection efficiency is more considered rather than accuracy to practical significance, especially on the potential abnormal behavior detection for high-speed and real-time network traffics. Nevertheless, both

**2.1 Efficiency Concerned Intrusion Detection**

In 1998, Lee and Stolfo [9] published a data mining approach for the intrusion detection, where they proposed a framework for the agent-based intrusion detection, and deployed data mining methods to extract detection rules. Afterwards many researchers definitely focused on the way of boosting the detection speed. For example, Sung[10] improved the detection speed by extracting the useful subset of attributes with ANN and SVM, and Srilatha[11] investigated the performance of Bayesian networks (BN) and Classification and Regression Trees (CART) to build lightweight IDS.

Actually, anomaly intrusion detection is a kind of complicated classification problem since there are usually too many attributes or features which may be redundant. So, attribute reduction or feature selection is the most popular method to improve detection efficiency by directly reducing the data attribute dimension[12]. There are several basic but still in-progressing ways of feature selection for anomaly intrusion detection. PCA is a widely used criteria to select features for intrusion detection, which can be usually incorporated with other soft computing models, such as neural networks[13], genetic algorithms[14], etc. PCA is a statistics-based method which is direct and effective, but it lacks much intelligence. Some novel methods are therefore proposed recently that employ diverse intelligent approaches to reduce features, such as fuzzy C means[15], mutual information definition[16], graph visualization technique[17], gradually removal method[18], etc. Although reducing attributes can obviously lower the time cost for detection, detection accuracy should also be guaranteed for real-world applications which will be discussed in the next subsection.

More importantly, as to handle the massive traffic data in network intrusion detection, two aspects of effort on efficiency improvement are further studied. One is to sample the data as a training dataset instead of considering the full dataset, for example the random data sampling method[19] was proposed to deal with the massive traffic, and the fuzzy C-means[20] was also employed for the purpose of selecting smaller number of training data. However, the biggest problem of sampling methods is that it may lose potentially useful data which are very distinguishable for detection[21]. Another important way is to parallelize the data processing procedure to achieve the computation improvement. With the development of the hardware, Giorgos firstly made efforts to improve the detection speed by applying graphics processors[22], followed by the improved works in paper [23] which focused on the scalability of analyzing large data except speed acceleration. At very current, MapReduce based cloud computing frameworks start to be studied to build efficient intrusion detection systems[24,25]. Actually, a general data mining toolkit using MapReduce has been integrated with Weka[26], and there also developed a new toolkit on Hadoop, Mahout[27], where most classical data mining algorithms are all parallelized and distributed. All these implementations show that cloud-based parallel computing is a promising solution to improve the efficiency of intrusion detection. It makes us believe that MapReduce could also be explored in our works to parallelize and distribute each step of data computing on model compression for intrusion detection.

**2.2 Accuracy Concerned Intrusion Detection**

As discussed above, intrusion data attributes are sometimes redundant, which is one intrinsic reason for attribute reduction. Nevertheless, attributes may also contain false correlation. As to reduce the false alarm rate, alert correlation algorithms are integrated with the existing feature selection methods to automatically generate a more concise feature set to improve the overall precision of intrusion detection[28,29,30]. Recently, some other machine learning-based adaptive methods are also proposed to purify the false alarm[31,32]. Other than feature selection, clustering has been widely used as hybrid machine learning approach to improve the detection rate[33].

Generally speaking, a hybrid approach for intrusion detection is involved between the clustering and classification models. Clustering is usually used as the first step to filter out unrepresentative data, while classification is used later for detection[34,35]. In fact, the data which cannot be clustered accurately could be regarded as noisy data or outliers. So, the final clustered data, also called representative data or exemplars, are possibly without the noisy data and could be better used to train the classifier to improve the detection rate[36]. As for clustering, k-means is the most popular solution for intrusion detection[35,37,38], while k-NN and SVM are two commonly accepted models for classification in this domain[35,36,38,39]. Moreover, improvement researches on k-NN and SVM are still continuingly active to raise the detection precision[38,40].

As a new study, recent researches show that a novel clustering method, affinity propagation[2], can well be used to generate the good representative data for intrusion detection training[41,42]. Also, the AP clustering method is already employed to well apply on network stream data processing to improve the detection correctness[43]. Although it can drive high accuracy, the clustering performance of AP when processing large training datasets is very frustrating. Therefore, research on parallelization of AP is also introduced recently, for example, GPU-based parallelization of AP has been freshly developed to speed up the clustering[44]. However, like the traditional parallel algorithms, GPU-based approaches still challenge some intrinsic problems such as shared communication, load balancing, etc[45]. So as aforementioned, implementing cloud computing based AP parallelization will achieve many merits such as the fault-tolerance of failure nodes, the ability to scale-up large number of nodes on commodity hardware, etc.

### 2.3 Both Concerned Intrusion Detection

From the related works discussed above, we can find the following two key technique trends on intrusion detection.
a. Clustering is very useful to find the outliers for intrusion detection[46], especially AP clustering is one promising method on exemplars selection for intrusion detection[41,42,43].
b. Cloud computing is a transparent framework to implement parallelization of data processing algorithms, especially MapReduce becomes a promising approach towards to scalable intrusion detection system[24,25,47].

Another observation is that most of the-state-of-art works on intrusion detection only focus on efficiency or accuracy. That means, feature selection and clustering are generally two independent research topics for efficiency and accuracy respectively, and there is few systematic and practical approach to guarantee both two.

Therefore, our works in this paper will initially focus on the following two aspects.
a. Propose a data compact model, named compressed model, which integrates both feature selection and AP clustering to be more concise and accurate. The compressed model is based on the view of systematic methodology, on which some classification model for intrusion detection can then be efficiently built.
b. Employ MapReduce to implement a full parallel compression framework which can be scalable in processing large training data. This makes the model compression both efficient and available, thus the classification-based intrusion detection using such compressed model is practical with high efficiency and accuracy.

## 3. Parallel and Scalable Model Compression

### 3.1 Using Training Data to Build Compressed Model

In this section, we will elaborate the details of our proposed compressed model. Conventionally, data collection, data analysis, model building and detection are four sequential steps for constructing the supervised or the semi-supervised anomaly detection system, where the model building step is our major concentration.

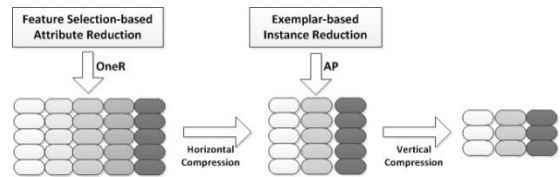

**Fig.2** Compressed model building with OneR as horizontal compression and AP as vertical compression

As illustrated in Fig.2, to build a compressed model, firstly we horizontally abstract the useful attributes from the training data that have been normalized, saying attribute reduction. Then, we vertically compress the training data to further select representative ones, saying instance reduction.

Prior to the discussion of the compressed model, we firstly introduce two datasets used throughout this paper. KDD99 dataset is a well-known intrusion detection evaluation dataset transformed from DARPA Intrusion Detection Evaluation dataset. Although KDD99 has been criticized for various reasons, it is still a benchmark for evaluating performance of intrusion detection. KDD99 is collected in a military network environment and it contains numerous simulated connections including normal ones and attacks. Each of the connections in the KDD99 has already been broken down into 41 attributes and well labeled as normal or a specific attack type. Here we simply classify all the instances into two types, normal and attack. Due to the huge volume of KDD99 and for experimental analysis purpose, we just use randomly selected 10,000 instances from KDD99 for training and another 90,000 instances for detection. Since KDD99 is an old dataset more than ten years ago which cannot identify the current network situation to some extent, we also use the CDMC2012 dataset in our experi-ments[1]. The real traffic data in CDMC2012 are collected from several types of honeypots and a mail server over 5 different networks inside and outside of Kyoto University. The dataset is composed of 14 features including label information which indicates whether each session is attack or not. Similarly, we classify the dataset into two categories, normal and attack, and we use the available training data to perform our experiments, where 12,872 instances are used for training and another 115,848 instances for detection.

Note that the whole training and testing data are always made up of a lot of instances. Each instance can be seen as a row, which consists of $m$ attributes, and can be expressed as $X_i = \{x_1, x_2, \ldots, x_j, \ldots, x_m\}$ ($1 \leq i \leq n, 1 \leq j \leq m$), where $X_i$ is the $i$-th instance of the whole dataset $X$ which totally contains $n$ instances. A mapping example of the attributes and their corresponding values in KDD99 dataset is listed in Table 1, and a similar mapping method is also employed in the later discussed dataset of CDMC2012.

**Table 1.** A mapping illustration of attributes and the corresponding values for KDD99

| Basic Attributes | | Content Attributes | | Traffic Attributes | |
|---|---|---|---|---|---|
| Name | Value | Name | Value | Name | Value |
| Duration | [0, 58329] | hot | [0, 101] | Count | [0, 511] |
| protocol_type | {TCP, UDP, ICMP} | logged_in | {0, 1} | serror_rate | [0.00, 1.00] |
| src_bytes | [0, 1379963888] | root_shell | {0, 1} | same_srv_rate | [0.00, 1.00] |
| wrong_fragment | [0, 3] | num_root | [0, 7468] | dst_host_count | [0, 255] |

3.1.1 Data Normalization

Since the training data are made up of a large number of instances and each instance has several attributes, a challenge about the training data is that the values of different attributes are distributed on disparate scales, which may cause a bias toward certain attributes over others. Here we give an example: consider two vectors with 3 attributes, {(0, 1200, 5), (1, 1000, 10)}. Taking the Euclidean distance for example, the squared distance between vectors will be $(0 - 1)^2 + (1200 - 1000)^2 + (5 - 10)^2$, which is decided heavily by the second attribute. To balance the contribution of every attribute in the similarity calculation, we first normalize the data to the scale of [0, 1] through the following formula (1).

$$X_i[j] = \frac{X_i[j] - X_{min}[j]}{X_{max}[j] - X_{min}[j]}. \tag{1}$$

Here $X_i[j]$ denotes the value of $j$th attribute of $i$th data instance. $X_{min}[j]$ is the smallest value of attribute $j$ among the dataset $X$, while $X_{max}[j]$ is the biggest one among all the data instances.

---

[1] CDMC2012 is the 3rd Cybersecurity Data Mining Competition satellited with ICONIP 2012, where a novel intrusion detection dataset is published as task 2. Note that two graduate students supervised by the first author have ever attended this competition and finally won the First place on task 2 with the highest detection accuracy 97.14%. You may please refer to http://www.csmining.org/cdmc2012/index.php?id=14 if interested.

3.1.2 Horizontal Compression

The horizontal compression, as the first step of building a compressed model, mainly explores the relations and correlations of the features within an instance, and extracts the key features. The horizontal compression is useful because in some complex classification fields the false correlative features may impede the process of detection. Most importantly, some features may be redundant since they may not play a role to distinguish one instance from others. Moreover, it is worth mentioning that excessive features may sharply slow down the detection. As like what we will introduce below, the clustering methods utilized in our experiments will calculate the distance between any two instances to measure their similarity. Given the high dimensionality resulting from numerous attributes in each instance, it is time-consuming to calculate such distance with all the attributes one by one. For this reason, it is meaningful to horizontally compress the training data no matter in the model building phase or in the detection process.

There are many ways to realize our horizontal compression idea. Here we choose OneR [48] as it is very easy to understand but with very high efficiency. We briefly introduce the principle of OneR by taking the *protocol_type* attribute of the KDD99 dataset as an example. The illustrated process of compressing the instances horizontally using OneR is shown in Fig.3.

| protocol_type | type |
|---|---|
| TCP | Normal |
| UDP | Attack |
| UDP | Attack |
| TCP | Normal |
| ICMP | Normal |
| UDP | Normal |
| TCP | Attack |
| ICMP | Normal |

| Summary | | type | |
|---|---|---|---|
| | | Normal | Attack |
| protocol_type | TCP | 2 | 1 |
| | UDP | 1 | 2 |
| | ICMP | 2 | 0 |

| protocol_type | prediction type | total error |
|---|---|---|
| TCP | Normal | |
| UDP | Attack | 1/4 |
| ICMP | Normal | |

**Fig. 3.** An illustration of how OneR works on *protocol_type* of the dataset from KDD99

OneR extracts the rules by examining the attributes one by one. The attribute in Fig.3 is *protocol_type*. First of all, OneR summarizes the quantitative relationship between the instance type (normal or attack) and the attribute value (TCP, UDP and ICMP). By traversing all the instances, it is concluded that there are two normal instances and one attack instance with the TCP *protocol_type* in Fig.3. Then OneR chooses the instance type (normal or attack) with a larger frequency as the prediction type for a certain attribute value. For example, since two normal instances (one attack instance) have the TCP value, the prediction type of the TCP is 'normal', which means if the instance has the *protocol_type* value TCP, we predict that this instance is a 'normal' one. Finally, OneR calculates the total error of every attribute, and chooses the attributes with relatively lower total errors from all attributes as the representative ones.

In our applications, as a default we finally select 12 out of 34 numerical attributes to represent an instance in KDD99 dataset, and 10 features out of total number of 14 after horizontal compression in CDMC2012.

### 3.1.3 Vertical Compression

Some training instances appear to be duplicate or similar, for example, they are probably the same packages and do the same business during a certain period of time. Therefore, vertical compression is called responsible for extracting a smaller set of representative training instances from a large scale dataset.

For the vertical compression purpose, we here employ the latest published but promising clustering approach, affinity propagation[2], which is actually a cornerstone through full of our idea. The reason to use the affinity propagation method instead of other traditional clustering methods, such as $k$-means, is straightforward that the affinity propagation can cluster the data without need to predefine a threshold $k$ of the expected number of clusters. Actually, in the case of the intrusion detection, we usually do not know how many clusters will be suitable for the training data.

Suppose that we have the knowledge about the appropriate number of clusters beforehand, the classical clustering method $k$-means, however, will also be conducted in our experiments for a comparison purpose. In order to make the parameter settings in our experiments understandable, we would introduce and discuss the $k$-means and the affinity propagation respectively in next section.

### 3.2 *K*-means and Affinity Propagation

### 3.2.1 Brief Descriptions on Methods

***K*-means.** Given a set of instances $X = \{X_1, X_2, \ldots, X_n\}$, where each instance is an attribute vector with real values. $k$-means aims to partition the $n$ instances into $k$ sets $(k < n)$ $S = \{S_1, S_2, \ldots, S_k\}$ so as to minimize the within-cluster sum of squares:

$$\sum_{i=1}^{k} \sum_{x_j \in s_i} \|X_j - \mu_i\|^2 . \quad (2)$$

Here $\mu_i$ is the mean point in $S_i$.

$K$-means firstly selects $k$ instances randomly as cluster centroids, and then it assigns each of the remaining instances to the cluster whose centroid is the most similar to this instance. After this, $k$-means refreshes all the clusters and makes the mean vector of the entire vectors within the cluster as the new centroid. $K$-means iteratively runs the procedure until the fitness function is convergent.

***Affinity Propagation.*** Affinity Propagation (AP) clusters instances by passing messages between data points iteratively. Define $X = \{X_1, X_2, \ldots, X_n\}$ as the instances to be clustered and let $d(X_i, X_j)$ denote the similarity or distance between instance $i$ and instance $j$. Finally, we should minimize the sum of the distances between instances and their exemplars. The fitness function is listed below:

$$E(c) = \sum_{i=1}^{n} S(X_i, X_{c(i)}) . \qquad (3)$$

Here $c(i)$ is the exemplar of instance $i$, and $S(X_i, X_j)$ is defined as below:

$$S(X_i, X_j) = \begin{cases} -d(X_i, X_j)^2 & if\ i \neq j \\ p & otherwise \end{cases} . \qquad (4)$$

Here $p$ stands for *preference* which is used to indicate how much an instance is likely to be chosen as an exemplar. Please note that an exemplar in AP is just like a representative instance of a cluster in *k*-means. The clustering procedure of AP seeks a good clustering result that can maximize the fitness function $E(c)$ by passing messages.

In summary, instead of requiring the number of clusters pre-specified in *k*-means, the preference in AP can affect the number of final clusters because the instance with larger preference is more likely to be chosen as an exemplar.

3.2.2 Discussions on performance

Although AP clustering is more suitable for vertical compression in our model than *K*-means, the time and space complexity of AP is both $O(N^2)$ while the *K*-means is $O(NK)$ where *K* is the specified initial number of clusters. That is to say, as running AP clustering on one single PC, computing the responsibility and availability matrix during iterations will easily run down when *N* increases, especially for the case that each data instance holds many attributes.

Fig.4 shows the quick performance degeneration of AP clustering on different subsets with different number of features selected from KDD99 data aforementioned compared to that of *K*-means, where the running termination of both algorithms is set to obtain the nearly same number of clusters for the sake of convincing comparison. Note that the different numbers of selected attributes are resulted from OneR with 20, 30 and 40 respectively. It can be seen from Fig.4 that, whatever the number of attributes used for clustering, the computation time cost of *K*-means is acceptable even if the training data items reaches to 10,000, while that of AP is with lower efficiency as the number of items increases. In fact, the running of AP is finally halted out of memory when the items are larger than 5,000 for the iterative computation on data matrix (Note that AP clustering definitely runs out of memory for 6000 training data instances in Fig.4.). Therefore, the general PC cannot at all afford the computation burden of AP clustering for large training dataset such as KDD99.

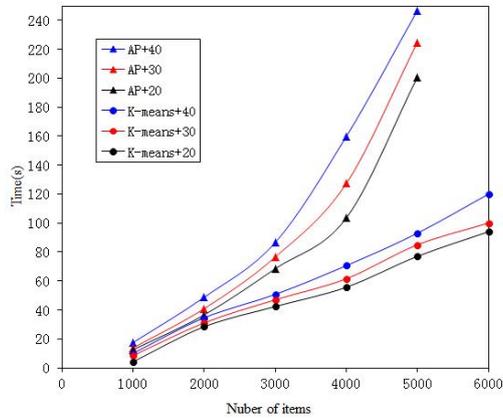

**Fig.4** Performances of *K*-means and AP clustering for different volumes of KDD99 training dataset with different number of selected attributes running on one PC equipped with AMD Phenom(tm) II N970 Quad-Core 2.20GHz CPU and 4GB RAM.

To solve this problem, we will propose a distributed and scalable solution for AP clustering on large training datasets using MapReduce. MapReduce is a distributed and parallel programming framework integrated in Apache Hadoop[49], which at current is a well-known popular cloud computing environment for large scale data processing. As a sequence, not only efficiency but also scalability for vertical compression conducted on large training data instances are expected when MapReduce is employed to parallelize each step of AP clustering.

## 3.3 Compression Using MapReduce

### 3.3.1 Implementations of Parallelization

A full solution of parallel computing using MapReduce for both horizontal and vertical compression is sketched as in Fig.5. OneR can be directly MapReduced by one-step, while as the procedure of AP is divided and MapReduced respectively and sequentially.

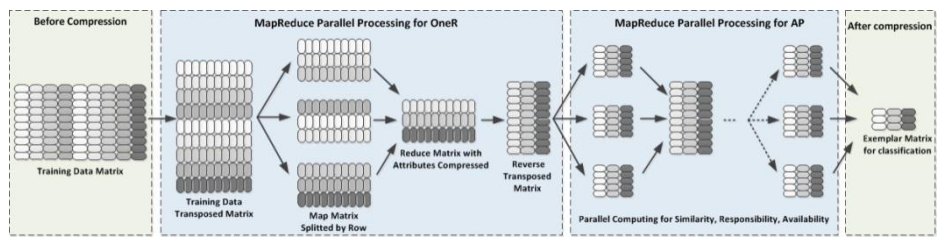

**Fig.5** The main skeleton of parallelization of model compression using MapReduce

Based on the work procedure of OneR, the parallelization could be easily realized if the error value of each attribute could be calculated on distributed nodes simultane-ously. Therefore, as shown in Fig.5, we firstly transpose the training data matrix to let each row of data matrix denote one values vector of each attribute, which can be taken as partitioned inputs for *Map* nodes. When each attribute's error value in OneR is parallelly generated by *Map* tasks on distributed nodes, the final selection of attributes can be conducted according to the summarization by one *Reduce* task.

Before looking into the parallelization of AP, let us review again the AP algorithm into deeper. Besides the similarity $s(i,k)$ defined between data point $i$ and $k$, responsibility $r(i,k)$ and availability $a(i,k)$ are other two key metrics in AP. The former shows the suitability of data point $k$ as the exemplar for data point $i$ after the $k$ competes all other point $k'$, while the later shows the contribution of data point $i$ to select $k$ as its exemplar after the $i$ evaluates all other point $i'$.

The iterative running steps of AP for $N$ total data points are described as below:

Step 1. Initialize the responsibility and availability matrix with all $r(i,k)$ and $a(i,k)$ being 0.
Step 2. Compute the similarity matrix $s(i,k)$.
Step 3. Update the responsibility $r(i,k)$ in term of the following rule:
$$r(i,k) = (1-\lambda) \times r(i,k) + \lambda \times (s(i,k) - \max_{k' \neq k}\{a(i,k') + s(i,k')\})$$
,where $\lambda$ is a damping parameter usually assigned with 0.8 according to experimental analysis[2].
Step 4. Update the availability $a(i,k)$ in term of the following rule:
$$a(i,k) = (1-\lambda) \times a(i,k) + \lambda \times (\min\{0, r(k,k) + \sum_{i' \notin \{i,k\}}\max\{0, r(i',k)\}\})$$
$$a(k,k) = (1-\lambda) \times a(k,k) + \lambda \times \sum_{i' \notin \{k\}}\max\{0, r(i',k)\}$$
Step 5. Select data point $k$ as clustering exemplar where $r(k,k)+a(k,k)>0$, and make data point $k$ as the exemplar for data point $i$ where $s(i,k)$ is the maximum one.
Step 6. Terminated if the iteration number exceeds a specific threshold or the exemplars are unchanged during a specific number of continued iterations, otherwise return step 3.

Observing the above steps of AP, we will then introduce the parallelization method for each step using MapReduce.

**Parallelization on Similarity Matrix Computation.** Euclidean distance is employed to compute the similarity under the following formula as for point $x$ and point $y$ with totally $n$ attributes:

$$s(x,y) = \sum_{k=1}^{n}(x_k - y_k)^2 = \sum_{k=1}^{n}x_k^2 + \sum_{k=1}^{n}y_k^2 - 2\sum_{k=1}^{n}x_k y_k$$

It is obvious that all the three terms for the training data matrix $A$ can be generated from the product of the matrix $A$ and its transposed one $A^T$, but here we will give a more tricky approach using MapReduce.

Assign $A$ is composed of $m$ instances with each instance having $n$ attributes:

$$A = \begin{bmatrix} a_{11}, a_{12}, ..., a_{1n} \\ a_{21}, a_{22}, ..., a_{2n} \\ ... \\ a_{m1}, a_{m2}, ..., a_{mn} \end{bmatrix}$$

The MapReduce procedure to compute the similarity matrix of $A$ is depicted in concise as following:

$$\begin{bmatrix} a_{11}, a_{12}, ..., a_{1n} \\ a_{21}, a_{22}, ..., a_{2n} \\ ... \\ a_{m1}, a_{m2}, ..., a_{mn} \end{bmatrix} \rightarrow \begin{bmatrix} a_{11}, a_{21}, ..., a_{m1} \\ a_{12}, a_{22}, ..., a_{m2} \\ ... \\ a_{1n}, a_{2n}, ..., a_{mn} \end{bmatrix} \rightarrow \begin{bmatrix} 1, a_{11} \times a_{11}, a_{11} \times a_{21}, ..., a_{11} \times a_{m1} \\ 2, 0\quad, a_{21} \times a_{21}, ..., a_{21} \times a_{m1} \\ ... \\ n, 0 \quad, 0 \quad, ..., a_{m1} \times a_{m1} \\ 1, a_{12} \times a_{12}, a_{12} \times a_{22}, ..., a_{12} \times a_{m2} \\ 2, 0 \quad, a_{22} \times a_{22}, ..., a_{22} \times a_{m2} \\ ... \\ n, 0 \quad, 0 \quad, ..., a_{mn} \times a_{mn} \end{bmatrix}$$

$$\rightarrow \begin{bmatrix} 1, a_{11} \times a_{11} + a_{12} \times a_{12} + ... + a_{1n} \times a_{1n}, a_{11} \times a_{21} + a_{12} \times a_{22} + ... + a_{1n} \times a_{2n}, ..., a_{11} \times a_{m1} + ... + a_{1n} \times a_{mn} \\ 2, 0 \quad, a_{21} \times a_{21} + a_{22} \times a_{22} + ... + a_{2n} \times a_{2n}, ..., a_{21} \times a_{m1} + ... + a_{2n} \times a_{mn} \\ ... \\ n, 0 \quad, 0 \quad, ..., a_{n1} \times a_{m1} + ... + a_{nn} \times a_{mn} \end{bmatrix}$$

As like for OneR, the matrix $A$ is firstly transposed to be $A^T$ in which each row is actually one vector made up of all values of one attribute. As for **Map** task, the respective products between each value (we call it key seed) and its backward values in each row are generated with assigning the column number of the key seed as **Key** and the product results list as **Values**. Here all the products between key seed and its forward values in each row are denoted with 0. Then, as for **Reduce** task, all the corresponding products in each row with the same **Key** value are combined to be aggregative addition terms, from which all the similarity items aforementioned can be simultaneously computed with ease.

**Parallelization on Responsibility and Availability Computation**. Note that if the damping parameter $\lambda$ is fixed, updating $r(i,k)$ is only related to the $i$th row of both availability and similarity matrix while $a(i,k)$ only to $k$th column of responsibility matrix. Hence, during MapReduce the **Map** task is only simply to assign each $i$th row as the **Key** value, and then the same $i$th row (with the **Key** value equaling $i$) of data point can be eventually shuffled to the same **Reduce** to update $r(i,k)$ and $a(i,k)$.

For efficiency purpose, we designed the following structure to present a data point:

*Point*
{
   int *x*;     //row value of matrix
   int *y*;     //column value of matrix
   double *s*; //value of $s(x, y)$
   double *r*; //value of $r(x, y)$
   double *a*; //value of $a(x,y)$
}

Such a point structure can store the similarity *s(x,y)*, responsibility *r(x,y)*, and availability *a(x,y)* of the point with *x*th row and *y*th column in training data matrix *N(x,y)*.

We firstly sequentialize each training data utilizing Point to be as input for MapReduce, then the above mentioned parallel computing procedure can be automatically implemented using the *Map* and *Reduce* program. Row-based MapReduce implementation for responsibility parallelization is pictured as Fig.6, while the availability is just with the similar process except for doing column-based *Map* and *Reduce*.

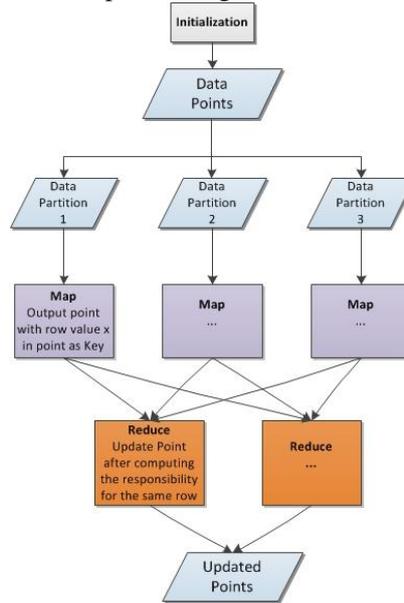

**Fig.6** Parallelization procedure for Responsibility updating in *Point* using MapReduce

**Parallelization on Exemplars Selection**. Because *a(k,k)* and *r(k,k)* are both included in one same *Point* structure, each *Map* node can independently compute which data points are exemplars. Therefore, computing parallelization for exemplar selection can be easily implemented by that *Map* task generates the corresponding exemplars and *Reduce* task directly combines all exemplars.

3.3.2 Evaluations on Time Cost and Scalability

As to horizontal compression using parallelization, it can ignore the time cost for that it only takes less than one second to compress totally 100,000 instances from KDD99 and 128,720 instances from CDMC2012 dataset.

The time cost for generating the compressed model vertically with AP and *k*-means is described comparably in Fig.7. It can be concluded that model compression may take relatively long time, but is considerably improved compared to that from single PC as show in Fig.4. In fact, the parallel running time for clustering 6000 training data points, even with 40 attributes, using MapReduce with 8 nodes only needs about 100 seconds. As to CDMC2012 dataset, however, the average time cost using the same parallelization platform for compressing model vertically is about 30 seconds.

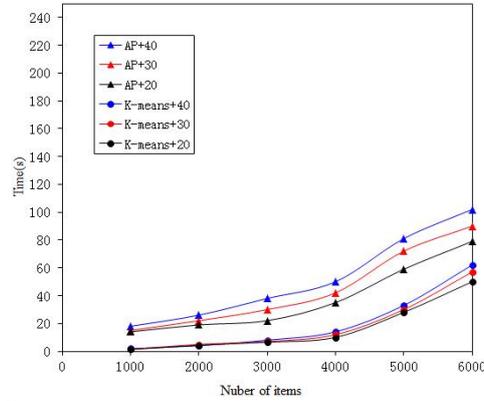

**Fig.7** Performances of *K*-means and AP clustering for different volumes of KDD99 training dataset with different number of selected attributes running on 8 nodes of MapReduce clusters with each node equipped with AMD Phenom(tm) II N970 Quad-Core 2.20GHz CPU and 4GB RAM.

Not like as single PC could not afford the computation burden of AP clustering for larger volume of training data with many attributes, the proposed MapReduce employed parallelization solution for AP is much more scalable for training data volume. As following the performance test results are presented in Fig.8 and Fig.9 respectively. Fig.8 shows the total speedup of computing power for AP-based data compressing by augmenting the number of nodes using MapReduce, while Fig.9 shows the specific scalability power for handling the increasing volume of training datasets with different number of nodes. We can clearly see that, from Fig.9, when 8 nodes employed in MapReduce, the time needed to handle 8,000 sample points is only around 3 times than that to 2,000 both for KDD99 and CDMC2012.

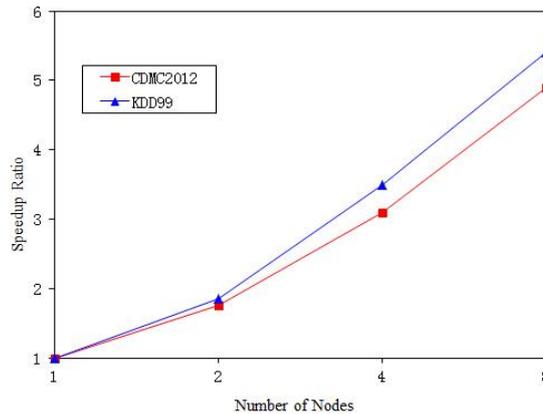

**Fig.8** Performance of AP clustering for different number of nodes using MapReduce for the given KDD99 and CDMC2012 dataset(also each node equipped with AMD Phenom(tm) II N970 Quad-Core 2.20GHz CPU and 4GB RAM)

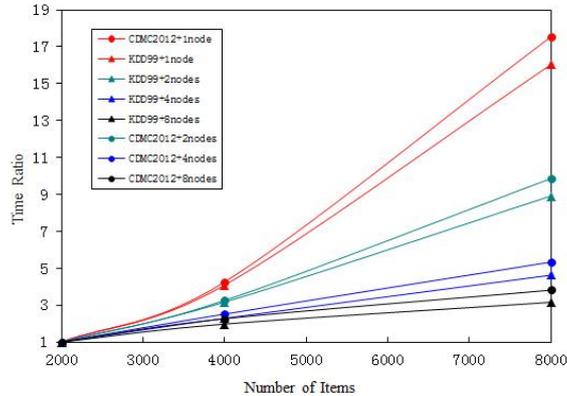

**Fig.9** Performance of AP clustering for different volumes of KDD99 and CDMC2012 training datasets using MapReduce with different number of nodes (also each node equipped with AMD Phenom(tm) II N970 Quad-Core 2.20GHz CPU and 4GB RAM)

Also, the figures in the Fig.7 give us a hint that the step of vertical compression can be done off-line to meet the requirement in the real-time environment, which applies to the horizontal compression as well. To be specific, we just need to provide the original training data to the MapReducer responsible for compressing the model, from which the generated compact model will be returned to the on-line part for the later classification modeling. From Fig.7, we can also observe that the horizontal compression with OneR can significantly save time on the vertical compression in most cases.

## 4. Classification Using Compressed Model

After the compressed model is built, two traditional classification methods are adopted in our experiments to evaluate the detection performance of the compressed model, namely KNN and SVM.

KNN (K Nearest Neighbor) is one of the most widely used classification methods in data mining. It finds $k$ nearest neighbors of a given instance among all the training data. There exist various ways of realizing KNN algorithm, and for the purpose of comparing traditional KNN with our improved KNN by utilizing distance matrix, a linear scan is employed in our experiments.

SVM (Support Vector Machine) is one of the recognized machine learning methods for classification, regression and other learning tasks. Here we would apply C-SVC (C–Support Vector Classification), one among the SVMs, to identify whether a package is abnormal or not. Unlike one-class SVM, which is a frequently used detection method in intrusion detection using the model built with normal class only, C-SVC is based on the model with both normal and abnormal instances which is exactly applicable for KDD99 and CDMC2012 datasets.

Since parameters of clustering methods and detection approaches can significantly influence the results, here we will explain the way we choose them.

**AP.** In AP, the only parameter that should be set is the ***preference***. In our experiments, for a comprehensive evaluation, we choose the preferences between the minimum and the maximum of the similarities to generate expected number of clusters that are separately distributed. The relationship between the number of exemplars generated with AP and the corresponding preference set is described in Table 2. Since the number of clusters generated by AP is decided by its preference, it is hardly possible to generate exactly the wanted number of exemplars. Thus, we endeavor to generate similar numbers of exemplars for the conditions with and without OneR respectively for a comparison purpose, but not exactly the same. Accordingly, it can be observed from Table 2 that the number of exemplars grows with the increment of the preference within expectation.

**Table 2.** The relationship between ***preference*** and the number of ***exemplar*** for KDD99 and CDMC2012 in AP

| (a) KDD99 | | | | (b) CDMC2012 | | | |
|---|---|---|---|---|---|---|---|
| *With OneR* | | *Without OneR* | | *With OneR* | | *Without OneR* | |
| Exemplar | Preference | Exemplar | Preference | Exemplar | Preference | Exemplar | Preference |
| 205 | -0.0482469 | 203 | -0.4156618 | 69 | -0.4384616 | 65 | -0.8939194 |
| 324 | -0.0170783 | 322 | -0.1557253 | 120 | -0.1488941 | 120 | -0.2527850 |
| 480 | -0.0072700 | 471 | -0.0672369 | 212 | -0.0434867 | 216 | -0.0729020 |
| 625 | -0.0036726 | 620 | -0.0368273 | 326 | -0.0160566 | 326 | -0.0260614 |
| 885 | -0.0015379 | 891 | -0.0156519 | 562 | -0.0041900 | 526 | -0.0084119 |
| 1029 | -0.0010654 | 1092 | -0.0090514 | 689 | -0.0023318 | 682 | -0.0040380 |
| 1218 | -0.0006846 | 1203 | -0.0067543 | 876 | -0.0011680 | 886 | -0.0020661 |
| 1537 | -0.0003841 | 1566 | -0.0028883 | 1215 | -0.0003477 | 1260 | -0.0004942 |
| 2110 | -0.0001585 | 2072 | -0.0011145 | 1618 | -0.0001059 | 1634 | -0.0001566 |
| 3044 | -0.0000385 | 3083 | -0.0002925 | 2056 | -0.0000116 | 2074 | -0.0000412 |
| 4043 | -0.0000077 | 4013 | -0.0001038 | 2512 | -0.0000006 | 2342 | -0.0000060 |

**K-means.** Since we use $k$-means here for the comparison purpose with AP, the parameter $k$ in the $k$-means should be set to the same with the number of exemplars generated by AP.

**KNN.** As to KNN, we set the parameter $k$ to be 1, which means that the type for each tested instance only depends on its nearest neighbor in the compressed model. Although this approach is rather simple, it has shown the high effectiveness practically.

**C-SVC.** As for C-SVC, since the number of attributes is quite small compared with the amount of instances, we choose to deploy a nonlinear kernel, namely Radial Basis Function kernel (RBF kernel), to map data to higher dimensional spaces. To better use the RBF kernel, one should determine two parameters: $C$ and $\gamma$. Since the test data are unknown in advance, we can only find proper parameters using foregone training data with the help of cross-validation. LibSVM [50] provides an automatic implementation on grid-searching $C$ and $\gamma$ using cross-validation, through which the best parameters for our training data can be concluded as following: for KDD99 dataset, the best $(C, \gamma) = (2048, 1.2207E - 4)$ when OneR is deployed while as $(C, \gamma) = (32768, 3.05176E - 5)$ without OneR, and for CDMC2012, dataset $(C, \gamma) = (8192, 3.05176E - 5)$ for both with and without OneR.

## 5. Application on Intrusion Detection

### 5.1 Detection Efficiency

The acceleration of detection is our major contribution in this paper. The test results about it are shown in Fig.10 and Fig.11 for both KDD99 and CDMC2012 with regard to our horizontal and vertical compression. Please note that baselines in Fig.10 and Fig.11, which do not employ the vertical compression but horizontal compression may be used, are drawn as well for a comparison purpose. By observing the two graphs, we can conclude that:

a. There is an obvious improvement of speed due to our horizontal compression when we compare every blue line with the dashed red line in each graph, where the maximal speed-up factor reaches to around 4 in KDD99 with KNN method.
b. Given horizontal compression, there is furthermore an obvious improvement of speed due to our vertical compression when we compare the red horizontal line with the red slash lines in each graph, where the maximal speed-up factor reaches to around 8 in KDD99 with KNN method.
c. It is reasonable to see from figures that the smaller amount of clusters (namely number of items as shown in these two figures) is, the shorter time it will take for detection for both KNN and SVM. One may tend to use smaller number of training instances to meet the need of real time if only the detection accuracy can be satisfied. But of course, as only if the MapReduce parallelization proposed in section 3 is employed for model compression, very large number of training instances can also be conducted to generate the clusters for detection usage.

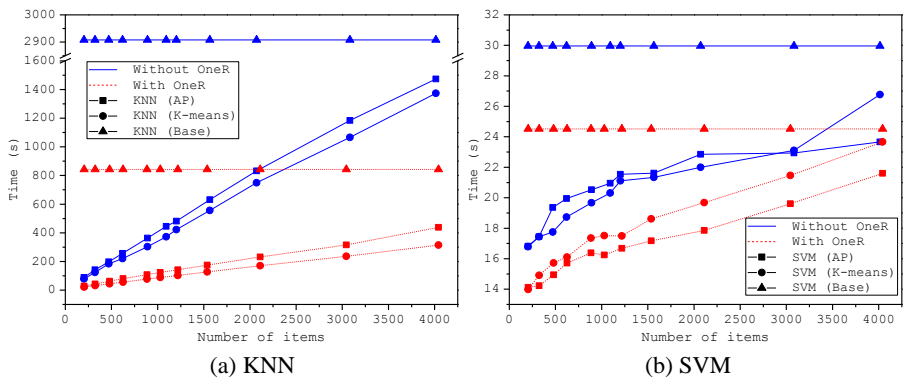

(a) KNN  (b) SVM

**Fig.10** Comparative results of detection time with KNN and SVM respectively for KDD99

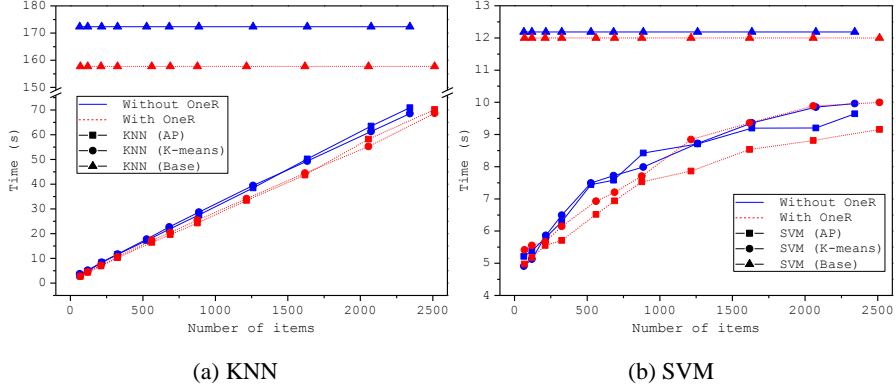

(a) KNN          (b) SVM

**Fig.11** Comparative results of detection time with KNN and SVM respectively for CDMC2012

Specifically, if we consider the blue baseline without any compression and the red resulting record with our full compressed model, the speedup ratio can reach to around 132 times (172.363 versus 1.3) for CDMC2012 and almost 184 times (2907.775 versus 15.718) for KDD99 when the simple KNN is directly used as the detection method.

### 5.2 Detection Accuracy

Although the resulting efficiency that has been empirically proved as aforementioned is our major focus in this paper, the detection accuracy is another important factor for a real-world detection application. However, we will concentrate on the difference of detection accuracy (with and without the compressed model) instead of the direct detection accuracy, because our compressed model is not proposed to contribute to the accuracy improvement, and the detection methods which are related to the direct detection accuracy are still the common ones. We would like to survey the difference with two standard measures here, namely recall and false positive rate. Recall is the ratio between the number of correctly detected anomalies and the total number of anomalies. False positive rate is the ratio between the number of data records from normal class that are misclassified as anomalies and the total number of data records from normal class. Table 3 and Table 4 record the recall and the false positive rate (not the relative difference) respectively for KDD99, and results of CDMC2012 are shown in Table 5 and Table 6. The last row of each table below is the benchmark without vertical compression(8,000 and 5,000 instances are selected from KDD99 and CDMC2012 respectively for experimental purpose).

**Table 3.** Recall with and without horizontal compression for KDD99

(a) With horizontal compression

| $R$ \ Num | KNN | | SVM | |
|---|---|---|---|---|
| | AP | K-means | AP | K-means |
| 205 | 98.6538 | 98.9963 | 97.1988 | 97.2399 |
| 324 | 98.0899 | 99.1644 | 97.2186 | 97.1364 |

(b) Without horizontal compression

| $R$ \ Num | KNN | | SVM | |
|---|---|---|---|---|
| | AP | K-means | AP | K-means |
| 203 | 99.4422 | 99.3577 | 97.5557 | 97.5329 |
| 322 | 99.5487 | 99.4513 | 97.9530 | 97.5291 |

| | | | | | | | | |
|---|---|---|---|---|---|---|---|---|
| 480 | 98.5229 | 99.2246 | 97.2194 | 97.1600 | 471 | 99.6157 | 99.4331 | 97.5702 | 97.5420 |
| 625 | 98.8182 | 99.1226 | 97.2148 | 97.1783 | 620 | 99.6355 | 99.6872 | 98.9635 | 98.6645 |
| 885 | 99.0708 | 99.0237 | 97.2109 | 97.1631 | 891 | 99.6545 | 99.6850 | 98.2315 | 99.2558 |
| 1029 | 98.8250 | 99.2344 | 97.1905 | 97.1783 | 1092 | 99.7085 | 99.6796 | 99.0160 | 99.0648 |
| 1218 | 98.9400 | 99.2786 | 97.1889 | 97.1958 | 1203 | 99.7192 | 99.6926 | 99.1089 | 99.1972 |
| 1537 | 99.1515 | 99.4003 | 97.1912 | 97.1943 | 1566 | 99.7177 | 99.7260 | 99.1850 | 99.1819 |
| 2110 | 98.9970 | 99.4551 | 97.1927 | 97.1927 | 2072 | 99.7162 | 99.7245 | 99.2550 | 99.2421 |
| 3044 | 99.3730 | 99.4909 | 97.1904 | 97.1889 | 3083 | 99.7177 | 99.7238 | 99.2306 | 99.2459 |
| 4043 | 99.3577 | 99.5099 | 97.1889 | 97.1882 | 4013 | 99.7184 | 99.7207 | 99.2558 | 99.2383 |
| 8000 | 99.5738 | | 97.2255 | | 8000 | 99.7215 | | 99.2436 | |

**Table 4.** False positive rate with and without horizontal compression for KDD99

(a) With horizontal compression  (b) Without horizontal compression

| FPR\Num | KNN | | SVM | | FPR\Num | KNN | | SVM | |
|---|---|---|---|---|---|---|---|---|---|
| | AP | K-means | AP | K-means | | AP | K-means | AP | K-means |
| 205 | 0.2699 | 0.7509 | 0.0196 | 0.0273 | 203 | 0.0552 | 0.0908 | 0.0338 | 0.0486 |
| 324 | 0.1857 | 0.5932 | 0.0219 | 0.0125 | 322 | 0.0374 | 0.0634 | 0.0279 | 0.0522 |
| 480 | 0.3138 | 0.6780 | 0.0225 | 0.0136 | 471 | 0.0996 | 0.1453 | 0.0285 | 0.0344 |
| 625 | 0.3073 | 0.5131 | 0.0225 | 0.0136 | 620 | 0.1311 | 0.2118 | 0.0409 | 0.0303 |
| 885 | 0.3779 | 0.4561 | 0.0225 | 0.0142 | 891 | 0.0919 | 0.2776 | 0.0350 | 0.0611 |
| 1029 | 0.2746 | 0.4146 | 0.0214 | 0.0196 | 1092 | 0.0913 | 0.2052 | 0.0480 | 0.0469 |
| 1218 | 0.2533 | 0.3553 | 0.0219 | 0.0237 | 1203 | 0.0747 | 0.2058 | 0.0623 | 0.0694 |
| 1537 | 0.3345 | 0.3262 | 0.0219 | 0.0231 | 1566 | 0.0629 | 0.1975 | 0.0712 | 0.0706 |
| 2110 | 0.2906 | 0.4217 | 0.0225 | 0.0231 | 2072 | 0.0581 | 0.1548 | 0.0605 | 0.0730 |
| 3044 | 0.2841 | 0.4336 | 0.0243 | 0.0255 | 3083 | 0.0581 | 0.1530 | 0.0700 | 0.0694 |
| 4043 | 0.3001 | 0.4288 | 0.0255 | 0.0249 | 4013 | 0.0581 | 0.0996 | 0.0706 | 0.0736 |
| 8000 | 0.3565 | | 0.0326 | | 8000 | 0.0991 | | 0.0937 | |

**Table 5.** Recall with and without horizontal compression for CDMC2012

(a) With horizontal compression  (b) Without horizontal compression

| R\Num | KNN | | SVM | | R\Num | KNN | | SVM | |
|---|---|---|---|---|---|---|---|---|---|
| | AP | K-means | AP | K-means | | AP | K-means | AP | K-means |
| 69 | 97.1653 | 95.2388 | 97.9342 | 93.2254 | 65 | 95.7145 | 95.3075 | 96.5955 | 94.6834 |
| 120 | 96.7909 | 95.8339 | 97.0098 | 93.4226 | 120 | 95.7308 | 78.0481 | 97.0785 | 87.6411 |
| 212 | 96.2301 | 96.9319 | 96.9320 | 94.3126 | 216 | 94.9855 | 79.8101 | 95.8412 | 90.6983 |
| 326 | 95.9425 | 96.0365 | 95.0868 | 93.4243 | 326 | 95.0054 | 78.9110 | 94.3343 | 93.9743 |
| 562 | 95.6024 | 89.9765 | 94.2782 | 93.8766 | 526 | 94.9602 | 79.1842 | 94.5333 | 94.2873 |
| 689 | 95.6458 | 90.1013 | 94.3886 | 93.9074 | 682 | 95.0217 | 76.0836 | 94.8860 | 94.7902 |
| 876 | 95.8755 | 90.2768 | 94.4935 | 93.9707 | 886 | 95.1954 | 87.6972 | 94.8444 | 94.8318 |
| 1215 | 95.9280 | 77.8365 | 94.7739 | 94.0503 | 1260 | 95.1592 | 90.2840 | 94.9367 | 94.7684 |
| 1618 | 95.9443 | 78.5709 | 94.7395 | 94.3614 | 1634 | 95.1592 | 88.6849 | 95.0000 | 94.8082 |
| 2056 | 95.8430 | 62.8817 | 94.7467 | 94.3813 | 2074 | 95.1429 | 62.7985 | 95.6078 | 94.9331 |
| 2512 | 95.8068 | 64.4736 | 94.5279 | 94.3849 | 2342 | 95.0742 | 62.8636 | 95.1302 | 94.9729 |
| 5000 | 96.0980 | | 94.3126 | | 5000 | 95.4016 | | 96.5955 | |

**Table 6.** False positive rate with and without horizontal compression for CDMC2012

(a) With horizontal compression  (b) Without horizontal compression

| FPR\Num | KNN | | SVM | | FPR\Num | KNN | | SVM | |
|---|---|---|---|---|---|---|---|---|---|
| | AP | K-means | AP | K-means | | AP | K-means | AP | K-means |
| 69 | 2.1786 | 2.1581 | 2.3042 | 2.2881 | 65 | 1.1338 | 1.1163 | 2.2560 | 2.1625 |
| 120 | 2.2735 | 2.0894 | 2.3042 | 2.2808 | 120 | 1.1455 | 0.7715 | 2.3042 | 1.9740 |
| 212 | 2.1815 | 2.0617 | 2.3042 | 2.2589 | 216 | 1.1163 | 0.8358 | 2.2911 | 1.9959 |
| 326 | 2.0967 | 2.0295 | 2.2881 | 2.1800 | 326 | 1.1178 | 0.0801 | 2.2677 | 2.1376 |
| 562 | 1.9798 | 1.7928 | 2.2750 | 2.2443 | 526 | 1.1192 | 0.8022 | 2.2662 | 2.2691 |
| 689 | 1.9842 | 1.8162 | 2.2750 | 2.2443 | 682 | 1.1222 | 0.7905 | 2.2691 | 2.2721 |
| 876 | 1.9258 | 1.7840 | 2.2443 | 2.2443 | 886 | 1.1178 | 1.0243 | 2.2735 | 2.2428 |
| 1215 | 1.9535 | 1.5079 | 2.2794 | 2.2443 | 1260 | 1.1178 | 1.0491 | 2.2428 | 2.2428 |
| 1618 | 1.8907 | 1.5108 | 2.2487 | 2.2443 | 1634 | 1.1178 | 1.0184 | 2.2443 | 2.2428 |
| 2056 | 1.8717 | 1.2946 | 2.2443 | 2.2443 | 2074 | 1.1178 | 0.5611 | 2.2458 | 2.2443 |
| 2512 | 1.8498 | 1.3004 | 2.2443 | 2.2443 | 2342 | 1.1002 | 0.5625 | 2.2443 | 2.2443 |
| 5000 | 1.8279 | | 2.2443 | | 5000 | 1.0929 | | 2.2531 | |

According to these tables, we can find that both of two measures (recall and false positive rate) from the *k*-means have a larger difference with that of the benchmark compared to the ones resulting from the affinity propagation (averaged at less than 1% for all the cases) which is actually the cornerstone of our work.

## 6. Conclusion and Future Work

We have proposed a compressed model for intrusion detection, including horizontal compression and vertical compression on training data. To be specific, the first endeavor in this paper is to horizontally select useful attributes of the training data through efficient method of OneR, and then vertically extract the representative data as exemplars from a larger dataset through the novel clustering method of AP. Based on the compressed model on training dataset, we finally studied KNN and SVM as two detection methods and empirically identified all parameters of AP for model compression and SVM for classification. Comprehensive experiments on intrusion detection datasets have been conducted to demonstrate the high performance resulting from the compressed model we proposed. In the best case, it runs 184 times faster than the traditional one without our model compression, and neither the recall (detection rate) nor the false positive rate sacrifices only in a very small range with less than 1% on average which can be tolerable or ignorable compared to the big efficiency improvement. Another important contribution is that we have implemented a MapReduce based parallelization framework for each step of the model compression procedure. Actually, we have made efforts to improve both the performance and scalability of compression on large volume of training data by MapReduce-based parallelization. Experimental analysis have been conducted to show that our model compression framework can handle more than 10,000 data points in minutes using 8 commodity PC nodes, while the memory of one general PC can only process not exceeding 5,000 instances using AP clustering.

The main merits of our works on intrusion detection are twofold. One is to initially introduce a compressed model of training data space for efficient intrusion detection which focuses on both efficiency and accuracy. The other is, to the best of our knowledge, the first effort of MapReduce based implementation to do the parallel and scalable compression on large amount of training data. However, there are also two open problems left to be solved in our approach, the ability of incremental compression processing and the performance refinement of MapReduce-based parallel computing framework.

As the future work, we will further develop our approach in two ways. One is to explore the study on incremental clustering ability of AP[51], as in order to efficiently handle the incremental clustering on network intrusion training data. Another, im-provement on the performance of MapReduce based parallel computing is a long-term goal, especially to study on the iteration mechanism employing some latest parallelization frameworks such as SPARK[52]. We hope our methodology proposed in this paper can finally be served as a general reference model to meet kinds of security mining challenge from big data systems rather than intrusion detection[53].

## Acknowledgements

This work was supported by the Natural Science Foundation of China with grant No.60113044, Zhejiang Provincial Science and Technology Project with grant No.2013C31G2020218. Also the authors would like to thank Dr.Rongsheng Gong and Prof.Samuel H. Huang from the Intelligent Systems Laboratory at University of Cincinnati for their valuable comments on this paper.